\documentclass{article}

     \usepackage[final, nonatbib]{neurips_2020}

\usepackage[utf8]{inputenc} 
\usepackage[T1]{fontenc}    
\usepackage{hyperref}       
\usepackage{url}            
\usepackage{booktabs}       
\usepackage{amsfonts}       
\usepackage{nicefrac}       
\usepackage{microtype}      
\usepackage{amsmath}
\usepackage{amssymb}
\usepackage{graphicx,wrapfig,lipsum}
\usepackage{soul}
\usepackage{subfig}
\usepackage{clipboard}
\usepackage{amsthm}

\usepackage{multirow}
\usepackage[table,xcdraw]{xcolor}

\newtheorem{theorem}{Theorem}

\newtheorem{proposition}{Proposition}

\title{Principal Neighbourhood Aggregation for Graph Nets}

\author{%
  Gabriele Corso\thanks{Equal contribution.}\\
  \small University of Cambridge\\
  \texttt{gc579@cam.ac.uk}
  \And
  Luca Cavalleri$^*$\\
  \small University of Cambridge\\
  \texttt{lc737@cam.ac.uk}
  \And
  Dominique Beaini\\
  \small InVivo AI\\
  \texttt{dominique@invivoai.com}
  \AND
  Pietro Li\`{o}\\
  \small University of Cambridge\\
  \texttt{pietro.lio@cst.cam.ac.uk}
  \And
  Petar Veli\v{c}kovi\'{c}\\
  \small DeepMind\\
  \texttt{petarv@google.com}
}

\begin{document}

\maketitle

\begin{abstract}
    Graph Neural Networks (GNNs) have been shown to be effective models for different predictive tasks on graph-structured data. Recent work on their expressive power has focused on isomorphism tasks and countable feature spaces. We extend this theoretical framework to include continuous features---which occur regularly in real-world input domains and within the hidden layers of GNNs---and we demonstrate the requirement for multiple aggregation functions in this context. Accordingly, we propose Principal Neighbourhood Aggregation (PNA), a novel architecture combining multiple aggregators with degree-scalers (which generalize the sum aggregator). Finally, we compare the capacity of different models to capture and exploit the graph structure via a novel benchmark containing multiple tasks taken from classical graph theory, alongside existing benchmarks from real-world domains, all of which demonstrate the strength of our model. With this work, we hope to steer some of the GNN research towards new aggregation methods which we believe are essential in the search for powerful and robust models.
    
\end{abstract}

\section{Introduction}

Graph Neural Networks (GNNs) have been an active research field for the last ten years with significant advancements in graph representation learning \cite{scarselli2009, Bronstein_2017, battaglia2018relational, hamilton2017representation}. However, it is difficult to understand the effectiveness of new GNNs due to the lack of standardized benchmarks \cite{dwivedi2020benchmarking} and of theoretical frameworks for their expressive power.

In fact, most work in this domain has focused on improving the GNN architectures on a set of graph benchmarks, without evaluating the capacity of their network to accurately characterize the graphs' structural properties. Only recently there have been significant studies on the expressive power of various GNN models \cite{xu2018gin, garg2020generalization, morris2018weisfeiler, murphy2019relational, sato2020survey}. However, these have mainly focused on the isomorphism task in domains with countable features spaces, and little work has been done on understanding their capacity to capture and exploit the underlying properties of the graph structure. 


We hypothesize that the aggregation layers of current GNNs are unable to extract enough information from the nodes' neighbourhoods in a single layer, which limits their expressive power and learning abilities. 

We first mathematically prove the need for multiple aggregators and propose a solution for the uncountable multiset injectivity problem introduced by \cite{xu2018gin}. Then, we propose the concept of degree-scalers as a generalization to the \textit{sum} aggregation, which allow the network to amplify or attenuate signals based on the degree of each node. Combining the above, we design the proposed \emph{Principal Neighbourhood Aggregation} (PNA) model and demonstrate empirically that multiple aggregation strategies improve the performance of the GNN.

Dehmamy \emph{et al.} \cite{dehmamy2019understanding} have also empirically found that using multiple aggregators (mean, sum and normalized mean), which extract similar statistics from the input message, improves the performance of GNNs on the task of graph moments. In contrast, our work extends the theoretical framework by deriving the necessity to use complementary aggregators. Accordingly, we propose the use of different statistical aggregations to allow each node to better understand the distribution of the messages it receives, and we generalize the \textit{mean} as the first of a set of possible \textit{n-moment} aggregators. In the setting of graph kernels, Cai \textit{et al.} \cite{cai2018simple} constructed a simple baseline using multiple aggregators. In the field of computer vision, Lee \textit{et al.} \cite{lee2016generalizing} empirically showed the benefits of combining \textit{mean} and \textit{max} pooling. These give us further confidence in the validity of our theoretical analysis.

We present a consistently well-performing and parameter efficient encode-process-decode architecture \cite{hamrick2018relational} for GNNs. This differs from traditional GNNs by allowing a variable number of convolutions with shared parameters. Using this model, we compare the performances of some of the most diffused models in the literature (GCN \cite{kipf2016gcn}, GAT \cite{velikovic2017gat}, GIN \cite{xu2018gin} and MPNN \cite{gilmer2017mpnn}) with our PNA.

Previous work on tasks taken from classical graph theory focuses on evaluating the performance of GNN models on a single task such as shortest paths \cite{velickovic2019neural, xu2019neural, Graves2016}, graph moments \cite{dehmamy2019understanding} or travelling salesman problem \cite{dwivedi2020benchmarking, joshi2019efficient}.
Instead, we took a different approach by developing a multi-task benchmark containing problems both on the node level and the graph level. Many of the tasks are based on dynamic programming algorithms and are, therefore, expected to be well suited for GNNs \cite{xu2019neural}. We believe this multi-task approach ensures that the GNNs are able to understand multiple properties simultaneously, which is fundamental for solving complex graph problems. Moreover, efficiently sharing parameters between the tasks suggests a deeper understanding of the structural features of the graphs. Furthermore, we explore the generalization ability of the networks by testing on graphs of larger sizes than those present in the training set.

To further demonstrate the performance of our model, we also run tests on recently proposed real-world GNN benchmark datasets \cite{dwivedi2020benchmarking, hu2020open} with tasks taken from molecular chemistry and computer vision. Results show the PNA outperforms the other models in the literature in most of the tasks hence further supporting our theoretical findings.

The code for all the aggregators, scalers, models (in PyTorch, DGL and PyTorch Geometric frameworks), architectures, multi-task dataset generation and real-world benchmarks is available \href{https://github.com/lukecavabarrett/pna}{here}.

\section{Principal Neighbourhood Aggregation}

In this section, we first explain the motivation behind using multiple aggregators concurrently. We then present the idea of degree-based scalers, linking to prior related work on GNN expressiveness. Finally, we detail the design of graph convolutional layers which leverage the proposed Principal Neighbourhood Aggregation.

\subsection{Proposed aggregators}

\begin{figure}[h]
\centering
\includegraphics[width=\textwidth]{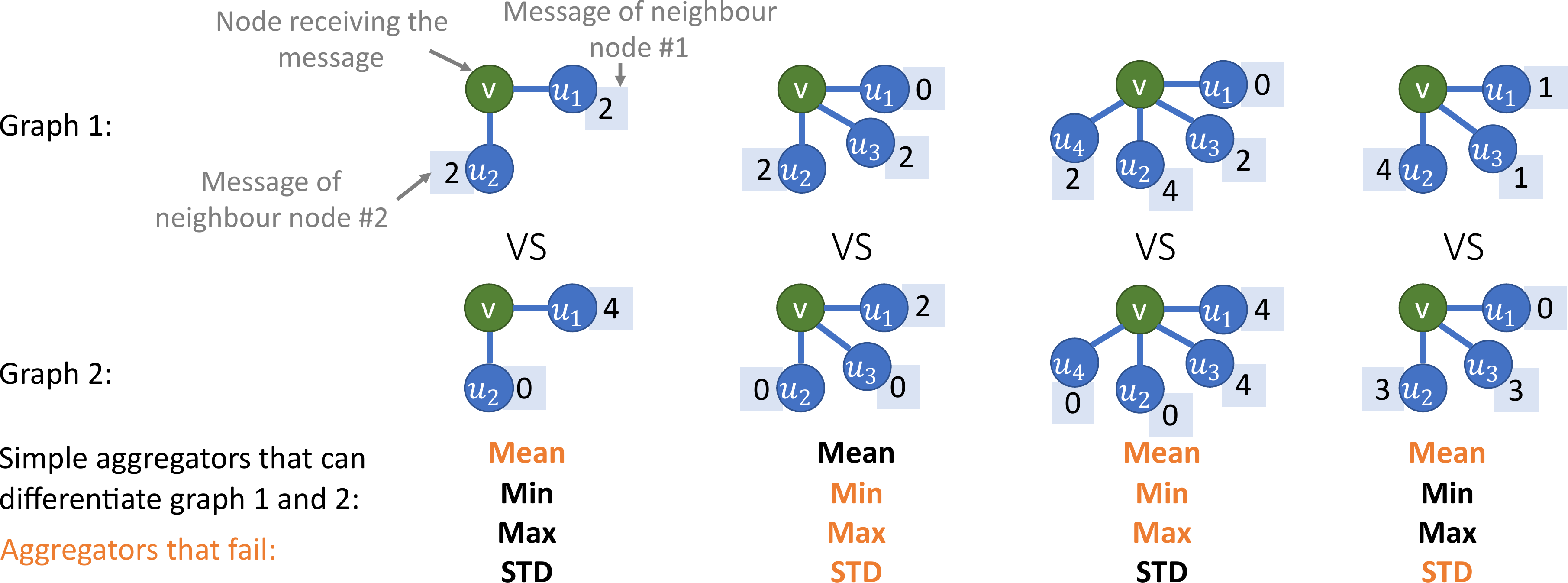}
\caption{Examples where, for a single GNN layer and continuous input feature spaces, some aggregators fail to differentiate between neighbourhood messages.} 
\label{fig:aggregators}
\end{figure}

Most work in the literature uses only a single aggregation method, with \textit{mean}, \textit{sum} and \textit{max} aggregators being the most used in the state-of-the-art models \cite{xu2018gin,kipf2016gcn,gilmer2017mpnn,velickovic2019neural}. In Figure \ref{fig:aggregators}, we observe how different aggregators fail to discriminate between different messages when using a single GNN layer.

We formalize our observations in the theorem below:

\begin{theorem}[Number of aggregators needed]
    \label{theorem:1}
    \Copy{theorem1}{
    \textit{In order to discriminate between multisets of size $n$ whose underlying set is $\mathbb{R}$, at least $n$ aggregators are needed.}
    }
\end{theorem}

\begin{proposition}[Moments of the multiset]
    \label{proposition:1}
    \Copy{proposition1}{
    \textit{The moments of a multiset (as defined in Equation \ref{eq:general_moments}) exhibit a valid example using $n$ aggregators.}}
\end{proposition}

We prove Theorem \ref{theorem:1} in Appendix \ref{app:proof_th1} and Proposition \ref{proposition:1} in Appendix \ref{app:proof_prop1}. Note that unlike Xu \emph{et al.} \cite{xu2018gin}, we consider a continuous input feature space; this better represents many real-world tasks where the observed values have uncertainty, and better models the latent node features within a neural network's representations. Continuous features make the space \underline{uncountable}, and void the injectivity proof of the \textit{sum} aggregation presented by Xu \emph{et al.} \cite{xu2018gin}. 

Hence, we redefine aggregators as continuous functions of multisets which compute a statistic on the neighbouring nodes, such as \textit{mean}, \textit{max} or \textit{standard deviation}. The continuity is important with continuous input spaces, as small variations in the input should result in small variations of the aggregators' output.


Theorem \ref{theorem:1} proves that the number of independent aggregators used is a limiting factor of the expressiveness of GNNs. To empirically demonstrate this, we leverage four aggregators, namely \textit{mean}, \textit{maximum}, \textit{minimum} and \textit{standard deviation}. Furthermore, we note that this can be extended to the \textit{normalized moment} aggregators, which allow advanced distribution information to be extracted whenever the degree of the nodes is high. 

The following paragraphs will describe the aggregators we leveraged in our architectures.

\paragraph{Mean aggregation $\mu(X^l)$}
The most common message aggregator in the literature, wherein each node computes a weighted average or sum of its incoming messages. Equation \ref{eq:mean} presents, on the left, the general mean equation, and, on the right, the direct neighbour formulation, where $X$ is any multiset, $X^l$ are the nodes' features at layer $l$, $N(i)$ is the neighbourhood of node $i$ and $d_i = | N(i) |$. For clarity we use $\mathbb{E}[f(X)]$ where $X$ is a multiset of size $d$ to be defined as $\mathbb{E}[f(X)] = \frac{1}{d} \sum_{x \in X} f(x)$.
\begin{equation}
    \label{eq:mean}
    \mu(X) = \mathbb{E}[X]
    \quad , \quad
    \mu_i(X^l) = \frac{1}{d_i} \sum_{j \in N(i)} X^l_{j}
\end{equation}

\paragraph{Maximum and minimum aggregations $\max(X^l), \; \min(X^l)$}
Also often used in literature, they are very useful for discrete tasks, for domains where credit assignment is important and when extrapolating to unseen distributions of graphs \cite{velickovic2019neural}. Alternatively, we present the softmax and softmin aggregators in Appendix \ref{app:aggregators}, which are differentiable and work for weighted graphs, but don't perform as well on our benchmarks.
\begin{equation}
    \label{eq:max}
    \text{max} _i(X^l) = \max_{j \in N(i)} X^l_j
    \quad , \quad
    \text{min} _i(X^l) = \min_{j \in N(i)} X^l_j
\end{equation}

\paragraph{Standard deviation aggregation $\sigma(X^l)$}
The standard deviation (STD or $\sigma$) is used to quantify the spread of neighbouring nodes features, such that a node can assess the diversity of the signals it receives. Equation \ref{eq:std} presents, on the left, the standard deviation formulation and, on the right, the STD of a graph-neighbourhood. \textit{ReLU} is the rectified linear unit used to avoid negative values caused by numerical errors and $\epsilon$ is a small positive number to ensure $\sigma$ is differentiable.
\begin{equation}
    \label{eq:std}
    \sigma(X) = \sqrt{\mathbb{E}[X^2] - \mathbb{E}[X]^2}
    \quad,\quad
\sigma_i(X^l) = \sqrt{ReLU \left( \mu_i({X^l}^2) - \mu_i{(X^l)}^2 \right) + \epsilon}
\end{equation}

\paragraph{Normalized moments aggregation $M_n(X^l)$}
The mean and standard deviation are the first and second normalized moments of the multiset ($n=1, n=2$). Additional moments, such as the skewness ($n=3$), the kurtosis ($n=4$), or higher moments, could be useful to better describe the neighbourhood. These become even more important when the degree of a node is high because four aggregators are insufficient to describe the neighbourhood accurately. As described in Appendix \ref{app:moments}, we choose the n\textsuperscript{th} root normalization, as presented in Equation \ref{eq:general_moments}, because it gives a statistic that scales linearly with the size of the individual elements (as the other aggregators); this gives the training adequate numerical stability. Once again we add an $\epsilon$ to the absolute value of the expectation before applying the n\textsuperscript{th} root for numerical stability of the gradient.
\begin{equation}
    \label{eq:general_moments}
    M_n(X) = \sqrt[n]{\mathbb{E} \left[(X - \mu)^n \right]}  \; \; , \; \; n > 1 
\end{equation}

\subsection{Degree-based scalers}
\label{sec:scalers}

We introduce scalers as functions of the number of messages being aggregated (usually the node degree), which are multiplied with the aggregated value to perform either an \emph{amplification} or an \emph{attenuation} of the incoming messages.

Xu \emph{et al.} \cite{xu2018gin} show that the use of \textit{mean} and \textit{max} aggregators by themselves fail to distinguish between neighbourhoods with identical features but with differing cardinalities, and the same applies to all the aggregators described above. They propose the use of the \textit{sum} aggregator to discriminate between such multisets. We generalise their approach by expressing the \textit{sum} aggregator as the composition of a \textit{mean} aggregator and a linear-degree amplifying scaler $S_{\text{amp}}(d) = d$.

\vspace{3pt}

\begin{theorem}[Injective functions on countable multisets]
    \label{theorem:2}
    \Copy{theorem2}{
    \textit{The \textit{mean} aggregation composed with any scaling linear to an injective function on the neighbourhood size can generate injective functions on bounded multisets of \underline{countable} elements. }
    }
\end{theorem}

We formalize and prove Theorem \ref{theorem:2} in Appendix \ref{app:proof_th2}; the results proven in \cite{xu2018gin} about the \textit{sum} aggregator become then a particular case of this theorem, and we can use any kind of injective scaler to discriminate between multisets of various sizes.

Recent work shows that summation aggregation doesn't generalize well to unseen graphs \cite{velickovic2019neural}, especially when larger. One reason is that a small change of the degree will cause the message and gradients to be amplified/attenuated exponentially (a linear amplification at each layer will cause an exponential amplification after multiple layers). Although there are different strategies to deal with this problem, we propose using a logarithmic amplification $S \propto \log(d+1)$ to reduce this effect. Note that the logarithm is injective for positive values, and $d$ is defined non-negative.  

Further motivation for using logarithmic scalers is to better describe the neighbourhood influence of a given node. Suppose we have a social network where nodes A, B and C have respectively 5 million, 1 million and 100 followers: on a linear scale, nodes B and C are closer than A and B; however, this does not accurately model their relative influence. This scenario exhibits how a logarithmic scale can discriminate better between messages received by \textit{influencer} and \textit{follower} nodes. 

We propose the logarithmic scaler $S_{\text{amp}}$ presented in Equation \ref{eq:S_amp}, where $\delta$ is a normalization parameter computed over the training set, and $d$ is the degree of the node receiving the message.
\begin{equation}
\label{eq:S_amp}
S_\text{amp}(d) =
\frac{\log(d + 1)}{\delta} \quad , \quad \delta = \frac{1}{|\text{train}|}\sum_{i \, \in \,  \text{train}}\log(d_i + 1)
\end{equation}

We further generalize this scaler in Equation \ref{eq:Scaler}, where $\alpha$ is a variable parameter that is negative for attenuation, positive for amplification or zero for no scaling. Other definitions of $S(d)$ can be used---such as a linear scaling---as long as the function is injective for $d>0$. 
\begin{equation}
\label{eq:Scaler}
S(d, \alpha) = \left(\frac{\log(d + 1)}{\delta} \right)^\alpha, \quad d>0, \quad -1 \leq \alpha \leq 1
\end{equation}

\subsection{Combined aggregation}

We combine the aggregators and scalers presented in previous sections obtaining the Principal Neighbourhood Aggregation (PNA). This is a general and flexible architecture, which in our tests we used with four neighbour-aggregations with three degree-scalers each, as summarized in Equation \ref{eq:PNA}. The aggregators are defined in Equations \ref{eq:mean}--\ref{eq:std}, while the scalers are defined in Equation \ref{eq:Scaler}, with $\otimes$ being the tensor product.
\begin{equation}
\label{eq:PNA}
\bigoplus = 
\underbrace{
\begin{bmatrix}
I \\ S(D, \alpha=1) \\ 
S(D, \alpha=-1)
\end{bmatrix} }_{\text{scalers}}
\otimes
\underbrace{\begin{bmatrix}
\mu \\ \sigma \\ \max \\ \min
\end{bmatrix}}_{\text{aggregators}}
\end{equation}

As mentioned earlier, higher degree graphs such as social networks could benefit from further aggregators (e.g. using the moments proposed in Equation \ref{eq:general_moments}). We insert the PNA operator within the framework of a message passing neural network \cite{gilmer2017mpnn}, obtaining the following GNN layer:
\begin{equation}
\label{eq:PNAmpnn}
X_i^{(t+1)} = 
U \left( X_i^{(t)}, 
\underset{(j,i) \in E}{\bigoplus} 
M \left( X_i^{(t)}, E_{j \rightarrow i}, X_j^{(t)} \right) \right) 
\end{equation}

where $E_{j \rightarrow i}$ is the feature (if present) of the edge $(j, i)$, $M$ and $U$ are neural networks (for our benchmarks, a linear layer was enough). $U$ reduces the size of the concatenated message (in space $\mathbb{R}^{13F}$) back to $\mathbb{R}^F$ where $F$ is the dimension of the hidden features in the network. As in the MPNN paper \cite{gilmer2017mpnn}, we employ multiple towers to improve computational complexity and generalization performance.

\begin{figure}[h]
\centering
\includegraphics[width=\textwidth]{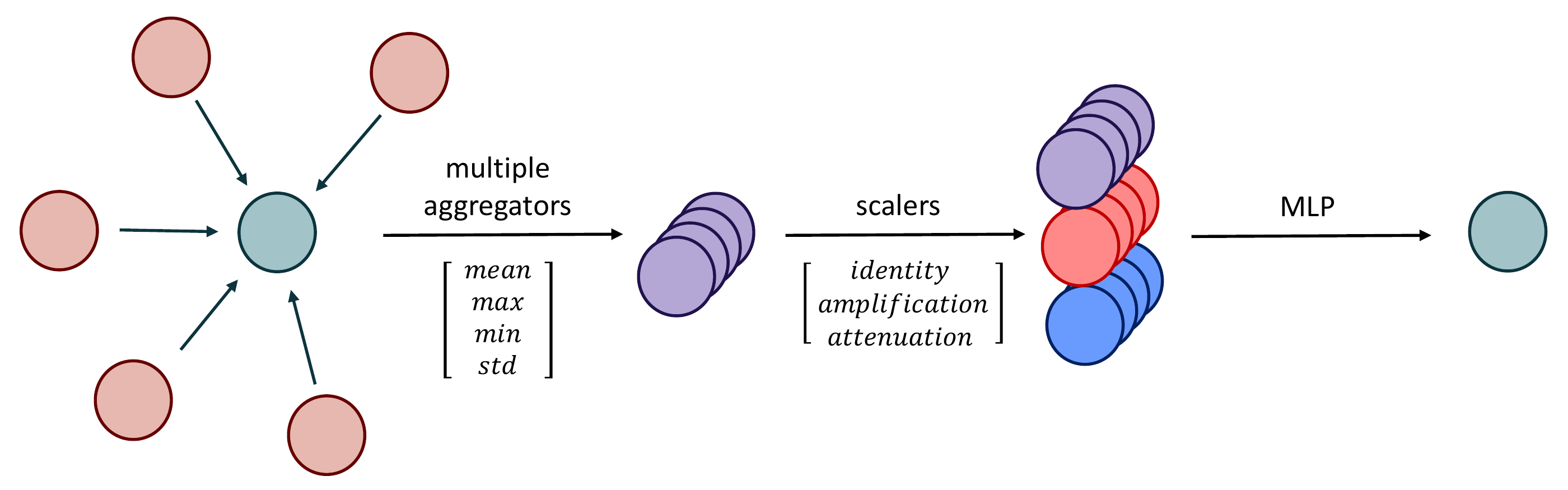}
\caption{Diagram for the Principal Neighbourhood Aggregation or PNA.} 
\label{fig:pna}
\end{figure}

Using twelve operations per kernel will require the usage of additional weights per input feature in the $U$ function, which could seem to be just quantitatively---not qualitatively---more powerful than an ordinary MPNN with a single aggregator \cite{gilmer2017mpnn}. However, the overall increase in parameters in the GNN model is modest and, as per our theoretical analysis above, a limiting factor of GNNs is likely their usage of a single aggregation.

This is comparable to convolutional neural networks (CNN) where a simple $3\times3$ convolutional kernel requires 9 weights per feature (1 weight per neighbour). Using a CNN with a single weight per $3\times3$ kernel will reduce the computational capacity since the feedforward network won't be able to compute derivatives or the Laplacian operator. Hence, it is intuitive that the GNNs should also require multiple weights per node, as previously demonstrated in Theorem \ref{theorem:1}. In Appendix \ref{app:parameters}, we will demonstrate this observation empirically, by running experiments on baseline models with larger dimensions of the hidden features (and, therefore, more parameters).

\section{Architecture} \label{sec:architecture}

We compare the performance of the PNA layer against some of the most popular models in the literature, namely GCN \cite{kipf2016gcn}, GAT \cite{velikovic2017gat}, GIN \cite{xu2018gin} and MPNN \cite{gilmer2017mpnn} on a common architecture. In Appendix \ref{app:convolutions}, we present the details of these graph convolutional layers.

\begin{figure}[h]
\centering
 \includegraphics[width=\textwidth]{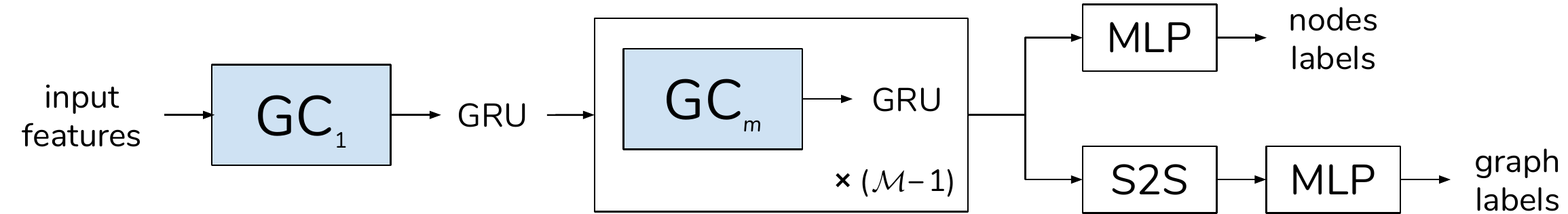}
 \caption{Layout of the architecture used. When comparing different models, the difference lies only in the type of graph convolution used in place of $GC_1$ and $GC_m$.}
\label{fig:architecture}
\end{figure}

For the multi-task experiments, we used an architecture, represented in Figure \ref{fig:architecture}, with $\mathcal{M}$ convolutions followed by three fully-connected layers for node labels and a set2set (S2S)\cite{vinyals2015order} readout function for graph labels. In particular, we want to highlight:

\textbf{Gated Recurrent Units (GRU)} \cite{cho2014learning} applied after the update function of each layer, as in \cite{gilmer2017mpnn, li2015gated}. Their ability to retain information from previous layers proved effective when increasing the number of convolutional layers $\mathcal{M}$.

\textbf{Weight sharing} in all the GNN layers but the first makes the architecture follow an encode-process-decode configuration \cite{battaglia2018relational, hamrick2018relational}. This is a strong prior which works well on all our experimental tasks, yields a parameter-efficient architecture, and allows the model to have a variable number $\mathcal{M}$ of layers.

\textbf{Variable depth} $\mathcal{M}$, decided at inference time (based on the size of the input graph and/or other heuristics), is important when using models over high variance graph distributions. In our experiments we have only used heuristics dependant on the number of nodes $N$ $(\mathcal{M} = f(N))$ and, for the architectures in the results below, we settled with $\mathcal{M}= \lfloor N / 2\rfloor$. It would be interesting to test heuristics based on properties of the graph, such as the diameter, or an adaptive computation time heuristic \cite{graves2016adaptive, spinelli2020adaptive} based on, for example, the convergence of the nodes features \cite{velickovic2019neural}. We leave these analyses to future work.

This architecture layout was chosen for its performance and parameter efficiency. We note that all architectural attempts yielded similar comparative performance of GNN layers and in Appendix \ref{app:std_architecture} we provide the results for a more \textit{standard} architecture.

\section{Multi-task benchmark}

The benchmark consists of classical graph theory tasks on artificially generated graphs.

\paragraph{Random graph generation}

Following previous work \cite{velickovic2019neural,you2019positionaware}, the benchmark contains undirected unweighted randomly generated graphs of a wide variety of types. In Appendix \ref{app:graph_generation}, we detail these types, and we describe the random toggling used to increase graph diversity. For the presented multi-task results, we used graphs of small sizes (15 to 50 nodes) as they were already sufficient to demonstrate clear differences between the models.

\paragraph{Multi-task graph properties}
    \label{sub:multitask}
In the multi-task benchmark, we consider three node labels and three graph labels based on standard graph theory problems. The node properties tasks are the single-source shortest-path lengths, the eccentricity and the Laplacian features ($L X$ where $L = (D - A)$ is the Laplacian matrix and $X$ the node feature vector). The graph properties tasks are whether the graph is connected, the diameter and the spectral radius.

\paragraph{Input features} As input features, the network is provided with two vectors of size $N$, a one-hot vector (representing the source for the shortest-path task) and a feature vector $X$ where each element is i.i.d. sampled as $X_i\sim\mathcal{U}[0,1]$. Apart from taking part in the Laplacian features task, this random feature vector also provides a \textit{unique identifier} for the nodes in other tasks. Similar strengthening via random features was also concurrently discovered by \cite{sato2020random}.
This allows for addressing some of the problems highlighted in \cite{garg2020generalization,chen2020can}; e.g. the task of whether a graph is connected could be performed by continually aggregating the maximum feature of the neighbourhood and then checking whether they are all equal in the readout.

\paragraph{Model training}
While having clear differences, these tasks also share related subroutines (such as graph traversals). While we do not take this sharing of subroutines as prior as in \cite{velickovic2019neural}, we expect models to pick up on these commonalities and efficiently share parameters between the tasks, which reinforce each other during the training.

We trained the models using the Adam optimizer for a maximum of 10,000 epochs, using early stopping with a patience of 1,000 epochs. Learning rates, weight decay, dropout and other hyper-parameters were tuned on the validation set. For each model, we run 10 training runs with different seeds and different hyper-parameters (but close to the tuned values) and report the five with least validation error.

\section{Results and discussion}

\subsection{Multi-task artificial benchmark}

\begin{figure}[h]
\centering

\subfloat[]{{\includegraphics[height=4.7cm]{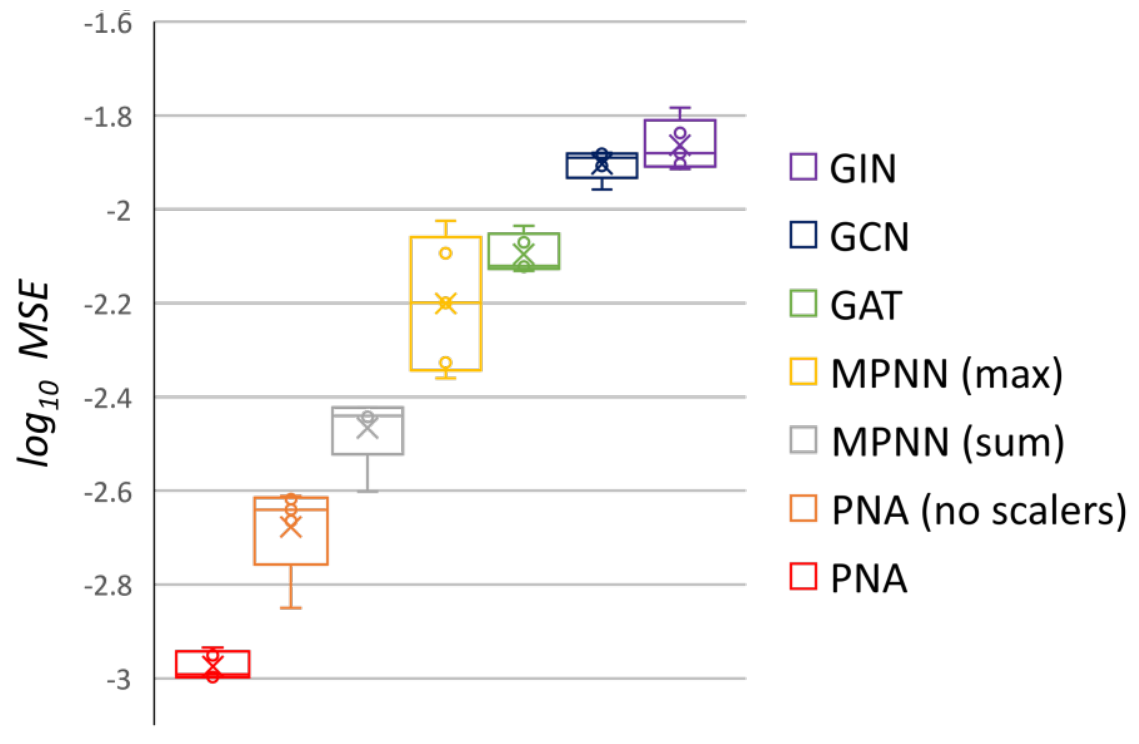} }
\label{fig:mse_bar_a}
}%
\subfloat[]{{\includegraphics[height=4.5cm]{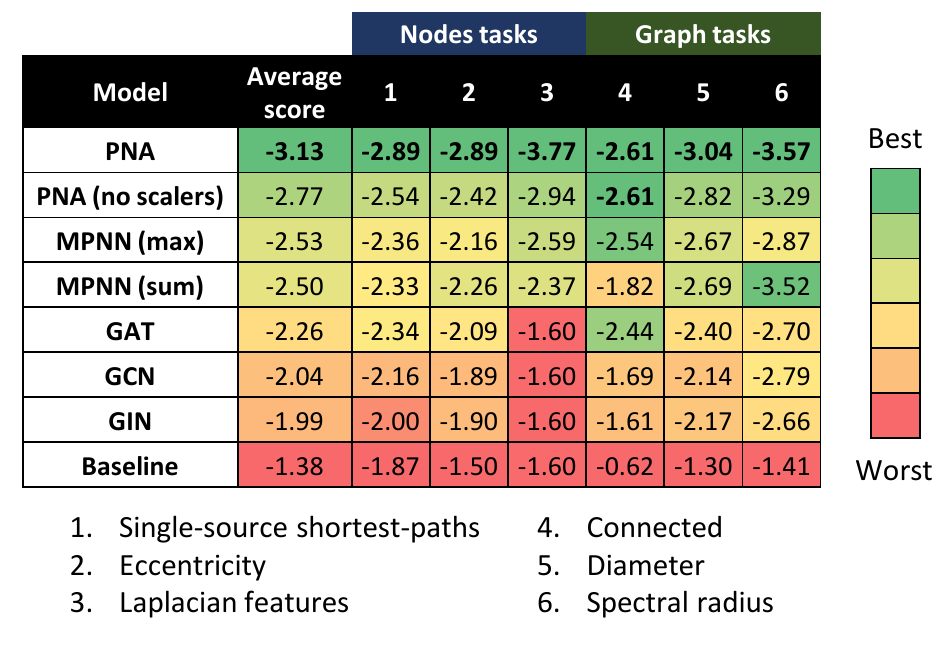} }
\label{fig:mse_bar_b}
}%

\caption{Multi-task benchmarks for different GNN models using the same architecture and various near-optimal hyper-parameters. (a) Distribution of the $\log_{10}$MSE errors for the top 5 performances of each model. (b) Mean $\log_{10}$MSE error for each task and their combined average. }
\label{fig:mse_bar}
\end{figure}

The multi-task results are presented in Figure \ref{fig:mse_bar_a}, where we observe that the proposed PNA model consistently outperforms state-of-the-art models, and in Figure \ref{fig:mse_bar_b}, where we note that the PNA performs better on all tasks. The \textit{baseline} represents the MSE from predicting the average of the training set for all tasks. 

The trend of these multi-task results follows and amplifies the difference in the average performances of the models when trained separately on the individual tasks. This suggests that the PNA model can better capture and exploit the common sub-units of these tasks. Appendix \ref{app:single_task} provides the average results of the models when trained on individual tasks. Moreover, PNA showed to perform the best on all architecture layouts that we attempted (see Appendix \ref{app:std_architecture}) and on all the various types of graphs (see Appendix \ref{app:graph_type}).



To demonstrate that the performance improvements of the PNA model are not due to the (relatively small) number of additional parameters it has compared to the other models (about 15\%), we ran tests on all the other models with latent size increased from 16 to 20 features. The results, presented in Appendix \ref{app:parameters}, suggest that even when these models are given 30\% more parameters than the PNA, they are qualitatively less capable of capturing the graph structure.

Finally, we explored the extrapolation of the models to larger graphs, in particular, we trained models on graphs of sizes between 15 and 25, validated between 25 and 30 and evaluate between 20 and 50. This task presents many challenges, two of the most significant are: firstly, unlike in \cite{velickovic2019neural} the models are not given any step-wise supervision or trained on easily extendable subroutines; secondly, the models have to cope with their architectures being augmented with further hidden layers than trained on, which can sometimes cause problems with rapidly increasing feature scales.

\begin{figure}[h]
\centering
\includegraphics[height=6.5cm]{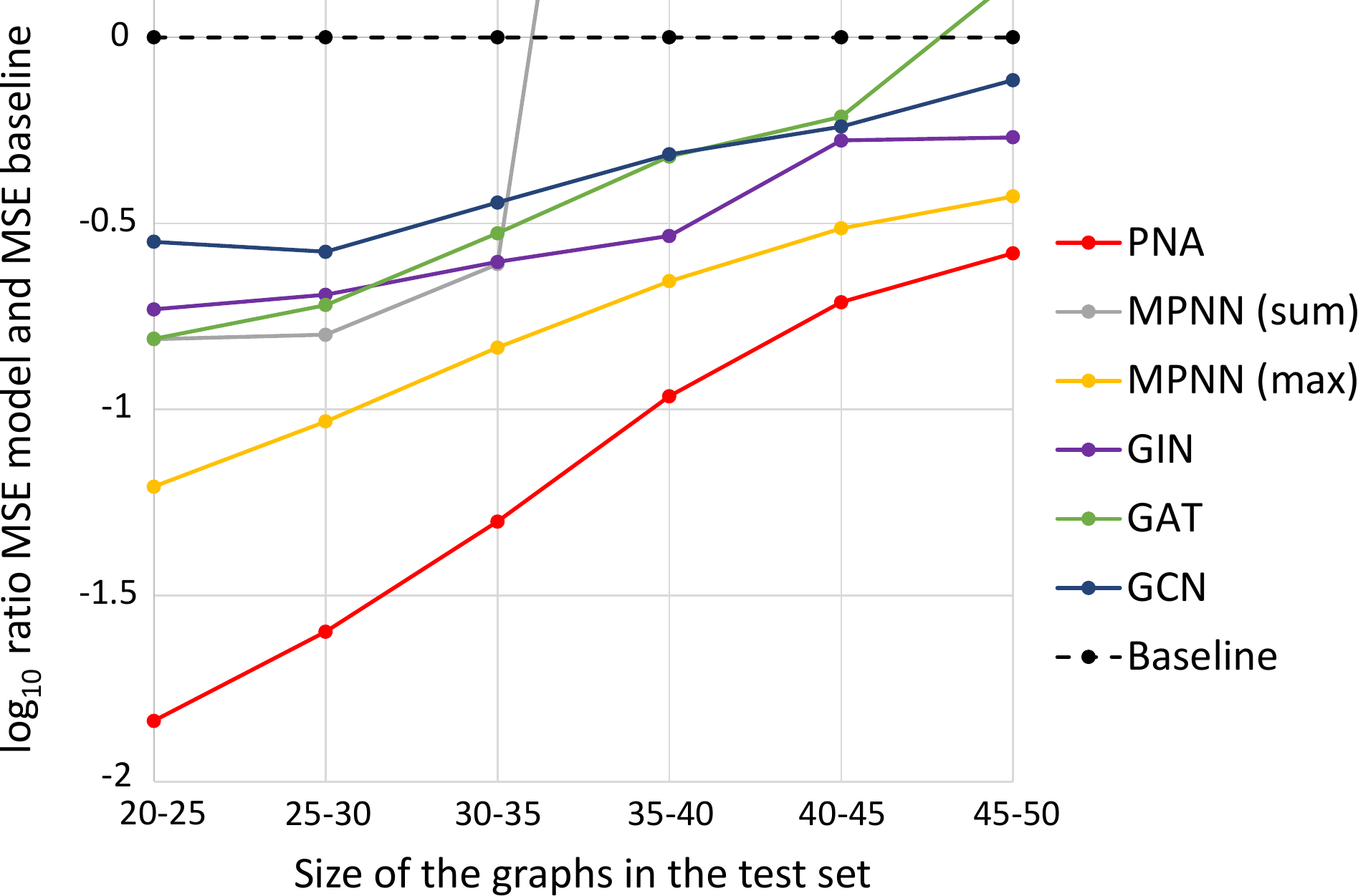}
\label{fig:mse_size_plot}
\caption{Multi-task $\log_{10}$ of the ratio of the MSE for different GNN models and the MSE of the baseline.}
\end{figure} 

Due to the aforementioned challenges, as expected, the performance of the models (as a proportion of the baseline performance) gradually worsens, with some of them having feature explosions. However, the PNA model keeps consistently outperforming all the other models on all graph sizes. Our results also follow the findings in \cite{velickovic2019neural}, i.e. that between single aggregators the \textit{max} tends to perform best when extrapolating to larger graphs. 

\subsection{Real-world benchmarks}

The recent works by Dwivedi \emph{et al.} \cite{dwivedi2020benchmarking} and Hu \emph{et al.} \cite{hu2020open} have shown problems with many benchmarks used for GNNs in recent years and proposed a new range of datasets across different artificial and real-world tasks. To test the capacity of the PNA model in real-world domains, we assessed it on their chemical (ZINC and MolHIV) and computer vision (CIFAR10 and MNIST) datasets.

To ensure a fair comparison of the different convolutional layers, we followed their method for training procedure (data splits, optimizer, etc.) and GNN structure (layers, normalization and approximate number of parameters). For the MolHIV dataset, we used the same GNN structure as in \cite{beaini2020directional}.

\begin{figure}[h]
\centering
\includegraphics[width=\textwidth]{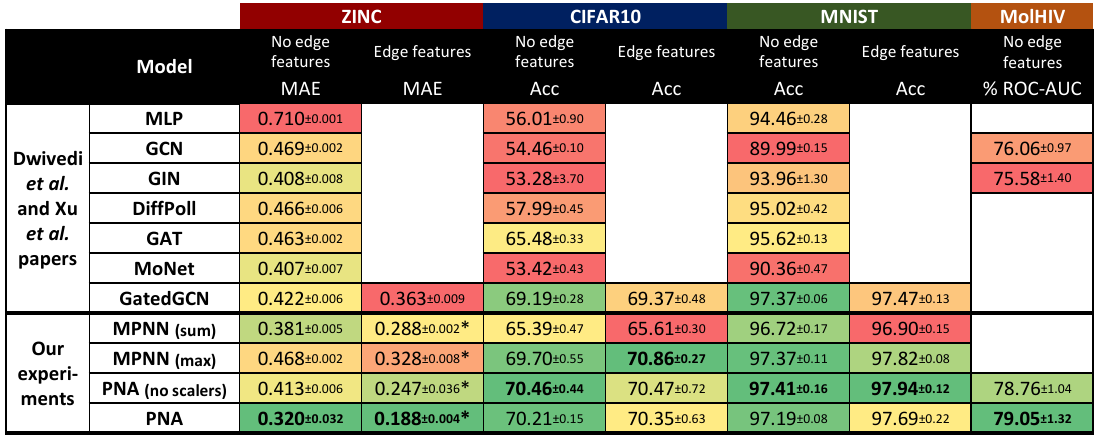}
\label{fig:real_world}
\caption{Results of the PNA and MPNN models in comparison with those analysed by Dwivedi \emph{et al.} and Xu \emph{et al.} (GCN\cite{kipf2016gcn},  GIN\cite{xu2018gin}, DiffPool\cite{ying2018hierarchical}, GAT\cite{velikovic2017gat}, MoNet\cite{monti2017moNet} and GatedGCN\cite{bresson2017gatedGCN}). * indicates the training was conducted with additional patience to ensure convergence.}
\end{figure} 

To better understand the results in the table, we need to take into account how graphs differ among the four datasets. In the chemical benchmarks, graphs are diverse and individual edges (bonds) can significantly impact the properties of the graphs (molecules). This contrasts with computer vision datasets made of graphs with a regular topology (every node has 8 edges) and where the graph structure of the representation is not crucial (the good performance of the MLP is evidence).

With this and our theoretical analysis in mind, it is understandable why the PNA has a strong performance in the chemical datasets, as it was designed to understand the graph structure and better retain neighbourhood information. At the same time, the version without scalers suffers from the fact it cannot distinguish between neighbourhoods of different size. Instead, in the computer vision datasets the average improvement of the PNA on SOTA was lower due to the smaller importance of the graph structure and the version of the PNA without scalers performs better as the constant degree of these graphs makes scalers redundant (and it is better to 'spend' parameters for larger hidden sizes).

\section{Conclusion}

We have extended the theoretical framework in which GNNs are analyzed to continuous features and proven the need for multiple aggregators in such circumstances. We also have generalized the $sum$ aggregation by presenting degree-scalers and proposed the use of a logarithmic scaling. Taking the above into consideration, we have presented a method, Principal Neighbourhood Aggregation, consisting of the composition of multiple aggregators and degree-scalers. With the goal of understanding the ability of GNNs to capture graph structures, we have proposed a novel multi-task benchmark and an encode-process-decode architecture for approaching it. Empirical results from synthetic and real-world domains support our theoretical evidence. We believe that our findings constitute a step towards establishing a hierarchy of models w.r.t. their expressive power, where the PNA model appears to outperform the prior art in GNN layer design.

\section*{Broader Impact}

Our work focuses mainly on theoretically analyzing the expressive power of Graph Neural Networks and can, therefore, play an indirect role in the (positive or negative) impacts that the field of graph representation learning might have on the domains where it will be applied.

More directly, our contribution in proving the limitations of existing GNNs on continuous feature spaces should help to provide an insight into their behaviour. We believe this is a significant result which might motivate future research aimed at overcoming such limitations, yielding more reliable models. However, we also recognize that, in the short-term, proofs of such weaknesses might spark mistrust against applications of these systems or steer adversarial attacks towards existing GNN architectures.

In an effort to overcome some of these short-term negative impacts and contribute to the search for more reliable models, we propose the Principal Neighbourhood Aggregation, a method that overcomes some of these theoretical limitations. Our tests demonstrate the higher capacity of the PNA compared to the prior art on both synthetic and real-world tasks; however, we recognize that our tests are not exhaustive and that our proofs do not allow for generating ``optimal'' aggregators for any task. As such, we do not rule out sub-optimal performance when applying the exact architecture proposed here to novel domains.

We propose the usage of aggregation functions, such as standard deviation and higher-order moments, and logarithmic scalers. To the best of our knowledge, these have not been used before in GNN literature. To further test their behaviour, we conducted out-of-distribution experiments, testing our models on graphs much larger than those in the training set. While the PNA model consistently outperformed other models and baselines, there was still a noticeable drop in performance. We therefore strongly encourage future work on analyzing the stability and efficacy of these novel aggregation methods on new domains and, in general, on finding GNN architectures that better generalize to graphs from unseen distributions, as this will be essential for the transition to industrial applications.

\section*{Acknowledgements}
The authors thank Saro Passaro for the valuable insights and discussion for the mathematical proofs.

\section*{Funding Disclosure}
Dominique Beaini is currently a Machine Learning Researcher at InVivo AI. Pietro Li\`{o} is a Full Professor at the Department of Computer Science and Technology of the University of Cambridge. Petar Veli\v{c}kovi\'{c} is a Research Scientist at DeepMind.

\bibliographystyle{unsrt}
\bibliography{citations}

\begin{thebibliography}{10}

\bibitem{scarselli2009}
F.~{Scarselli}, M.~{Gori}, A.~C. {Tsoi}, M.~{Hagenbuchner}, and
  G.~{Monfardini}.
\newblock The graph neural network model.
\newblock {\em IEEE Transactions on Neural Networks}, 20(1):61--80, 2009.

\bibitem{Bronstein_2017}
Michael~M. Bronstein, Joan Bruna, Yann LeCun, Arthur Szlam, and Pierre
  Vandergheynst.
\newblock Geometric deep learning: Going beyond euclidean data.
\newblock {\em IEEE Signal Processing Magazine}, 34(4):18–42, Jul 2017.

\bibitem{battaglia2018relational}
Peter~W Battaglia, Jessica~B Hamrick, Victor Bapst, Alvaro Sanchez-Gonzalez,
  Vinicius Zambaldi, Mateusz Malinowski, Andrea Tacchetti, David Raposo, Adam
  Santoro, Ryan Faulkner, et~al.
\newblock Relational inductive biases, deep learning, and graph networks.
\newblock {\em arXiv preprint arXiv:1806.01261}, 2018.

\bibitem{hamilton2017representation}
William~L Hamilton, Rex Ying, and Jure Leskovec.
\newblock Representation learning on graphs: Methods and applications.
\newblock {\em arXiv preprint arXiv:1709.05584}, 2017.

\bibitem{dwivedi2020benchmarking}
Vijay~Prakash Dwivedi, Chaitanya~K Joshi, Thomas Laurent, Yoshua Bengio, and
  Xavier Bresson.
\newblock Benchmarking graph neural networks.
\newblock {\em arXiv preprint arXiv:2003.00982}, 2020.

\bibitem{xu2018gin}
Keyulu Xu, Weihua Hu, Jure Leskovec, and Stefanie Jegelka.
\newblock How powerful are graph neural networks?
\newblock {\em arXiv preprint arXiv:1810.00826}, 2018.

\bibitem{garg2020generalization}
Vikas~K Garg, Stefanie Jegelka, and Tommi Jaakkola.
\newblock Generalization and representational limits of graph neural networks.
\newblock {\em arXiv preprint arXiv:2002.06157}, 2020.

\bibitem{morris2018weisfeiler}
Christopher Morris, Martin Ritzert, Matthias Fey, William~L Hamilton, Jan~Eric
  Lenssen, Gaurav Rattan, and Martin Grohe.
\newblock Weisfeiler and leman go neural: Higher-order graph neural networks.
\newblock In {\em Proceedings of the AAAI Conference on Artificial
  Intelligence}, volume~33, pages 4602--4609, 2019.

\bibitem{murphy2019relational}
Ryan~L Murphy, Balasubramaniam Srinivasan, Vinayak Rao, and Bruno Ribeiro.
\newblock Relational pooling for graph representations.
\newblock {\em arXiv preprint arXiv:1903.02541}, 2019.

\bibitem{sato2020survey}
Ryoma Sato.
\newblock A survey on the expressive power of graph neural networks.
\newblock {\em arXiv preprint arXiv:2003.04078}, 2020.

\bibitem{dehmamy2019understanding}
Nima Dehmamy, Albert-L{\'a}szl{\'o} Barab{\'a}si, and Rose Yu.
\newblock Understanding the representation power of graph neural networks in
  learning graph topology.
\newblock In {\em Advances in Neural Information Processing Systems}, pages
  15387--15397, 2019.

\bibitem{cai2018simple}
Chen Cai and Yusu Wang.
\newblock A simple yet effective baseline for non-attributed graph
  classification.
\newblock {\em arXiv preprint arXiv:1811.03508}, 2018.

\bibitem{lee2016generalizing}
Chen-Yu Lee, Patrick~W Gallagher, and Zhuowen Tu.
\newblock Generalizing pooling functions in convolutional neural networks:
  Mixed, gated, and tree.
\newblock In {\em Artificial intelligence and statistics}, pages 464--472,
  2016.

\bibitem{hamrick2018relational}
Jessica~B Hamrick, Kelsey~R Allen, Victor Bapst, Tina Zhu, Kevin~R McKee,
  Joshua~B Tenenbaum, and Peter~W Battaglia.
\newblock Relational inductive bias for physical construction in humans and
  machines.
\newblock {\em arXiv preprint arXiv:1806.01203}, 2018.

\bibitem{kipf2016gcn}
Thomas~N Kipf and Max Welling.
\newblock Semi-supervised classification with graph convolutional networks.
\newblock {\em arXiv preprint arXiv:1609.02907}, 2016.

\bibitem{velikovic2017gat}
Petar Veli{\v{c}}kovi{\'c}, Guillem Cucurull, Arantxa Casanova, Adriana Romero,
  Pietro Lio, and Yoshua Bengio.
\newblock Graph attention networks.
\newblock {\em arXiv preprint arXiv:1710.10903}, 2017.

\bibitem{gilmer2017mpnn}
Justin Gilmer, Samuel~S Schoenholz, Patrick~F Riley, Oriol Vinyals, and
  George~E Dahl.
\newblock Neural message passing for quantum chemistry.
\newblock In {\em Proceedings of the 34th International Conference on Machine
  Learning-Volume 70}, pages 1263--1272. JMLR. org, 2017.

\bibitem{velickovic2019neural}
Petar Veli{\v{c}}kovi{\'c}, Rex Ying, Matilde Padovano, Raia Hadsell, and
  Charles Blundell.
\newblock Neural execution of graph algorithms.
\newblock {\em arXiv preprint arXiv:1910.10593}, 2019.

\bibitem{xu2019neural}
Keyulu Xu, Jingling Li, Mozhi Zhang, Simon~S Du, Ken-ichi Kawarabayashi, and
  Stefanie Jegelka.
\newblock What can neural networks reason about?
\newblock {\em arXiv preprint arXiv:1905.13211}, 2019.

\bibitem{Graves2016}
Alex Graves, Greg Wayne, Malcolm Reynolds, Tim Harley, Ivo Danihelka, Agnieszka
  Grabska-Barwinska, Sergio~G{\'o}mez Colmenarejo, Edward Grefenstette, Tiago
  Ramalho, John Agapiou, Adri{\`a}~Puigdom{\`e}nech Badia, Karl~Moritz Hermann,
  Yori Zwols, Georg Ostrovski, Adam Cain, Helen King, Christopher Summerfield,
  Phil Blunsom, Koray Kavukcuoglu, and Demis Hassabis.
\newblock Hybrid computing using a neural network with dynamic external memory.
\newblock {\em Nature}, 538(7626):471--476, 2016.

\bibitem{joshi2019efficient}
Chaitanya~K Joshi, Thomas Laurent, and Xavier Bresson.
\newblock An efficient graph convolutional network technique for the travelling
  salesman problem.
\newblock {\em arXiv preprint arXiv:1906.01227}, 2019.

\bibitem{hu2020open}
Weihua Hu, Matthias Fey, Marinka Zitnik, Yuxiao Dong, Hongyu Ren, Bowen Liu,
  Michele Catasta, and Jure Leskovec.
\newblock Open graph benchmark: Datasets for machine learning on graphs.
\newblock {\em arXiv preprint arXiv:2005.00687}, 2020.

\bibitem{vinyals2015order}
Oriol Vinyals, Samy Bengio, and Manjunath Kudlur.
\newblock Order matters: Sequence to sequence for sets.
\newblock {\em arXiv preprint arXiv:1511.06391}, 2015.

\bibitem{cho2014learning}
Kyunghyun Cho, Bart Van~Merri{\"e}nboer, Caglar Gulcehre, Dzmitry Bahdanau,
  Fethi Bougares, Holger Schwenk, and Yoshua Bengio.
\newblock Learning phrase representations using rnn encoder-decoder for
  statistical machine translation.
\newblock {\em arXiv preprint arXiv:1406.1078}, 2014.

\bibitem{li2015gated}
Yujia Li, Daniel Tarlow, Marc Brockschmidt, and Richard Zemel.
\newblock Gated graph sequence neural networks.
\newblock {\em arXiv preprint arXiv:1511.05493}, 2015.

\bibitem{graves2016adaptive}
Alex Graves.
\newblock Adaptive computation time for recurrent neural networks.
\newblock {\em arXiv preprint arXiv:1603.08983}, 2016.

\bibitem{spinelli2020adaptive}
Indro Spinelli, Simone Scardapane, and Aurelio Uncini.
\newblock Adaptive propagation graph convolutional network.
\newblock {\em arXiv preprint arXiv:2002.10306}, 2020.

\bibitem{you2019positionaware}
Jiaxuan You, Rex Ying, and Jure Leskovec.
\newblock Position-aware graph neural networks.
\newblock {\em arXiv preprint arXiv:1906.04817}, 2019.

\bibitem{sato2020random}
Ryoma Sato, Makoto Yamada, and Hisashi Kashima.
\newblock Random features strengthen graph neural networks.
\newblock {\em arXiv preprint arXiv:2002.03155}, 2020.

\bibitem{chen2020can}
Zhengdao Chen, Lei Chen, Soledad Villar, and Joan Bruna.
\newblock Can graph neural networks count substructures?
\newblock {\em arXiv preprint arXiv:2002.04025}, 2020.

\bibitem{beaini2020directional}
Dominique Beaini, Saro Passaro, Vincent L{\'e}tourneau, William~L Hamilton,
  Gabriele Corso, and Pietro Li{\`o}.
\newblock Directional graph networks.
\newblock {\em arXiv preprint arXiv:2010.02863}, 2020.

\bibitem{ying2018hierarchical}
Zhitao Ying, Jiaxuan You, Christopher Morris, Xiang Ren, Will Hamilton, and
  Jure Leskovec.
\newblock Hierarchical graph representation learning with differentiable
  pooling.
\newblock In {\em Advances in neural information processing systems}, pages
  4800--4810, 2018.

\bibitem{monti2017moNet}
Federico Monti, Davide Boscaini, Jonathan Masci, Emanuele Rodola, Jan Svoboda,
  and Michael~M Bronstein.
\newblock Geometric deep learning on graphs and manifolds using mixture model
  cnns.
\newblock In {\em Proceedings of the IEEE Conference on Computer Vision and
  Pattern Recognition}, pages 5115--5124, 2017.

\bibitem{bresson2017gatedGCN}
Xavier Bresson and Thomas Laurent.
\newblock Residual gated graph convnets.
\newblock {\em arXiv preprint arXiv:1711.07553}, 2017.

\bibitem{rotman_introduction_1988}
Joseph~J. Rotman.
\newblock {\em An Introduction to Algebraic Topology}, volume 119 of {\em
  Graduate Texts in Mathematics}.
\newblock Springer New York.

\bibitem{borsuk}
Karol Borsuk.
\newblock Drei sätze über die n-dimensionale euklidische sphäre.
\newblock {\em Fundamenta Mathematicae}, (20):177–190, 1933.

\bibitem{newton}
Richard~P. Stanley.
\newblock {\em Enumerative Combinatorics Volume 2}.
\newblock Cambridge Studies in Advanced Mathematics no. 62. Cambridge
  University Press, Cambridge, 2001.

\bibitem{vieta}
M~Hazewinkel.
\newblock {\em Encyclopaedia of mathematics. Volume 9, STO-ZYG.}
\newblock Encyclopaedia of mathematics ; vol 9: STO-ZYG. Kluwer Academic,
  Dordecht, 1988.

\bibitem{erdos1960}
P.~Erd\H{o}s and A~R\'{e}nyi.
\newblock On the evolution of random graphs.
\newblock pages 17--61, 1960.

\bibitem{albert2002}
R\'{e}ka Albert and Albert-L\'{a}szl\'{o} Barab\'{a}si.
\newblock Statistical mechanics of complex networks.
\newblock {\em Reviews of Modern Physics}, 74(1):47–97, Jan 2002.

\bibitem{watts1999caveman}
Duncan~J. Watts.
\newblock Networks, dynamics, and the small‐world phenomenon.
\newblock {\em American Journal of Sociology}, 105(2):493--527, 1999.

\end{thebibliography}

\newpage

\appendix

\section{Proof for Theorem \ref{theorem:1} (\nameref{theorem:1})}
\label{app:proof_th1}
\Paste{theorem1}
\begin{proof}
Let $S$ be the $n$-dimensional subspace of $\mathbb{R}^n$ formed by all tuples $(x_1, \, x_2, \ldots , x_n)$ such that $x_1 \leq x_2 \leq \ldots \leq x_n$, and notice how $S$ is the collection of the aforementioned multisets. We defined an aggregator as a continuous function from multisets to reals, which corresponds to a continuous function $g : S \rightarrow \mathbb{R}$.

Assume by contradiction that it is possible to discriminate between all the multisets of size $n$ using only $n-1$ aggregators, viz. $g_1,g_2, \ldots , g_{n-1}$.

Define $f \! \! : \! \! S \! \! \rightarrow \! \! \mathbb{R}^{n-1}$ to be the function mapping each multiset $X$ to its output vector $(g_1(X), \, g_2(X),  \, \ldots \, , g_{n-1}(X))$.
Since $g_1,g_2, \ldots , g_{n-1}$ are continuous, so is $f$, and, since we assumed these aggregators are able to discriminate between all the multisets, $f$ is injective.

As $S$ is a $n$-dimensional Euclidean subspace, it is possible to define a $(n-1)$-sphere $C^{n-1}$ entirely contained within it, i.e. $C^{n-1} \subseteq S$. According to Borsuk–Ulam theorem \cite{rotman_introduction_1988, borsuk}, there are two distinct (in particular, non-zero and antipodal) points $\vec{x}_1,\vec{x}_2 \in C^{n-1}$ satisfying $f(\vec{x}_1) = f(\vec{x}_2)$, showing $f$ not to be injective; hence the required contradiction.
\end{proof}

\textit{Note:} \textit{$n$ aggregators are actually sufficient.}
A simple example is to use $g_1,g_2, \ldots , g_n$ where $g_k(X) =$ the $k$-th smallest item in $X$. It's clear to see that the multiset whose elements are $g_1(X), \, g_2(X),  \, \ldots \, , g_n(X)$ is $X$, which can hence be uniquely determined by the aggregators.

\section{Proof for Proposition \ref{proposition:1} (\nameref{proposition:1})}
\label{app:proof_prop1}
\Paste{proposition1}

\begin{proof}
Since $n \geq 1$, and the first aggregator is \textit{mean}, we know $\mu$.
Let $X = \{x_1,x_2,\ldots,x_n\}$ be the multiset to be found, and define $R = \{r_1=x_1-\mu,\; r_2=x_2-\mu,\; \ldots,\; r_n=x_n-\mu\}$.


Notice how $\sum {r_i}^1 = 0$, and for $1 < k \leq n$ we have $\sum {r_i}^k = n\; M_k(X)^k$, i.e. all the symmetric power sums $p_k = \sum {r_i}^k$ ($k \leq n$) are uniquely determined by the moments.

Additionally, $e_k$, the elementary symmetric sums of $R$, i.e. the sum of the products of all the sub-multisets of size $k$ ($1\leq k\leq n$), are determined as follow: 

$e_1$, the sum of all elements, is equal to $p_1$;  $e_2 $,  the sum of the products of all pairs in $R$, is $ \left(e_1p_1 - p_2 \right) / 2$; $e_3$, the sum of the products of all triplets, is $\left(e_2p_1 - e_1p_2 + p_3 \right) / 3 $, and so on. Notice how $e_1,e_2,\ldots,e_n$ can be computed using the following recursive formula \cite{newton}:
\begin{align*}
    \sum_{1\le i_1 < i_2 < \cdots < i_k\le n} \left(\prod_{j = 1}^k r_{i_j}\right)=e_k=\frac{1}{k}\sum_{j=1}^k (-1)^{j-1}e_{k-j}p_j
    \quad , \quad
    e_0 = 1
\end{align*}

Consider polynomial $P(x) = \Pi (x - r_i) $, i.e. the unique polynomial of degree $n$ with leading coefficient 1  whose roots are $R$. This defines $A$, the coefficients of $P$, i.e. the real numbers $a_0,a_1,\ldots,a_{n-1}$ for which  $P(x) = x^n + a_{n-1} x^{n-1} + \ldots + a_1 x + a_0$.
Using Vieta's formulas \cite{vieta}:
\begin{align*}
    \sum_{1\le i_1 < i_2 < \cdots < i_k\le n} \left(\prod_{j = 1}^k r_{i_j}\right)=(-1)^k\frac{a_{n-k}}{a_n}
\end{align*}
we obtain
\begin{align*}
    e_k &= (-1)^k\frac{a_{n-k}}{a_n} \\ 
        &= (-1)^k a_{n-k} && \text{recall} \;a_n = 1\\
     \therefore \: a_i &= (-1)^{n+i}e_{n+i} && \text{letting}\; k = n+i \; \text{and rearranging}
\end{align*}

Hence $A$ is uniquely determined, and so is $P$, being its coefficients a valid definition of it.
By the fundamental theorem of algebra, $P$ has $n$ (possibly repeated) roots, which are the elements of $R$, hence uniquely determining the latter.

Finally, $X$ can be easily determined adding $\mu$ to each element of $R$.
\end{proof}
\textit{Note:} \textit{the proof above assumes the knowledge of $n$.}
In the case that $n$ is variable (as in GNNs), and so we have multisets of \underline{\smash{up to}} $n$ elements, an extra aggregator will be needed. An example of such aggregator is the \textit{mean} multiplied by any injective scaler which would allow the degree of the node to be inferred.

\section{Proof for Theorem \ref{theorem:2} (\nameref{theorem:2})}
\label{app:proof_th2}
\Paste{theorem2}
\begin{proof}

Let $\chi$ be the countable input feature space from which the elements of the multisets are taken and $X$ an arbitrary multiset. Since $\chi$ is countable and the cardinality of multisets is bounded, let $Z:\chi \rightarrow \mathbb{N}^+$ be an injection from $\chi$ to natural numbers, and $N \in \mathbb{N}$ such that $|X|+1<N$ for all $X$.

Let's define an injective function $s$, and without loss of generality, assume $s(0), s(1), \ldots , s(N) > 0$ (otherwise for the rest of the proof consider $s$ as $s'(i) = s(i) - \min_{j \in [0,N]} s(j) + \epsilon$ which is positive for all $i \in [0,N]$). $s(|X|)$ can only take value in $\{ s(0), s(1), \ldots , s(N) \}$, therefore let us define $\gamma = \min \left \{ \frac{s(i)}{s(j)} \, \mid \, i,j \in [0, N], \: s(i) \geq s(j)\right \}$. Since $s$ is injective, $s(i) \neq s(j)$ for $i \neq j$, which implies $\gamma > 1$.
 
Let $K > \frac{1}{\gamma - 1}$ be a positive real number and consider $f(x)=N^{-Z(x)} + K$.

$\forall x \in \chi , Z(x) \in [1,N] \Rightarrow N^{-Z(x)} \in [0,1] \Rightarrow \: f(x) \in [K,K \! + \! 1] \,$, so $\, \mathbb{E}_{\, x \in X} [f(x)] \in [K, K \! + \! 1]$. 

We proceed to show that the cardinality of $X$ can be uniquely determined, and $X$ itself can be determined as well, by showing that exist an injection $h$ over the multisets.

Let us $h$ as a function that scales the mean of $f$ by an injective function of the cardinality:
$$h(X) = s(|X|) \: \mathbb{E}_{\, x \in X} [f(x)]$$

We want show that the value of $|X|$ can be uniquely inferred from the value of $h(X)$. Assume by contradiction $\exists \, X', X''$ multisets of size at most $N$ such that $|X'| \neq |X''|$ but $h(X') = h(X'')$; since $s$ is injective $s(|X'|) \neq s(|X''|)$, without loss of generality let $s(|X'|) > s(|X''|)$, then:
$$ s(|X''|) (K+1) \geq s(|X''|) \, \mathbb{E}_{\, x \in X''} [f(x)] = h(X'') = h(X') = s(|X'|) \, \mathbb{E}_{\, x \in X'} [f(x)] \geq s(|X'|) \, K$$
$$ \implies \: K \leq \frac{1}{\frac{s(|X'|)}{s(|X''|)} - 1} \leq \frac{1}{\gamma - 1}$$
which is a contradiction. So it is impossible for the size of a multiset $X$ to be ambiguous from the value of $h(X)$.

Let us define $d$ as the function mapping $h(X)$ to $|X|$.

$$h'(X) = \sum_{x \in X} N^{-Z(x)} = \frac{h(X) |X|}{s(|X|)} - K |X| = \frac{h(X) d(h(X))}{s(d(h(X)))} - K d(h(X))$$

Considering the $Z(j)$-th digit $i$ after the decimal point in the base $N$ representation of $h'(X)$, it can be inferred that $X$ contains $i$ elements $j$, and, so, all the elements in $X$ can be determined; hence $h$ is injective over the multisets in $X$.
\end{proof}
\textit{Note:} this proof is a generalization of the one by \textit{Xu et al.} \cite{xu2018gin} on the \textit{sum} aggregator.

\section{Normalized moments aggregation }
\label{app:moments}

The main motivation for choosing the n\textsuperscript{th} root normalization for the moments is numerical stability. In fact, one property of our version is that it scales linearly with $L$, for uniformly distributed random variables $U[0, L]$, as do other aggregators such as mean, max and min (std is a particular case). Other common formulations of the moments such as those in Equation \ref{eq:other_moments} scale respectively as the n\textsuperscript{th} power and constantly with $L$. This difference causes numerical instability when combined in the same layer.
\begin{equation}
    \label{eq:other_moments}
    M_n(X) = \mathbb{E} \left[(X - \mu)^n \right] \quad \quad
    M_n(X) = \frac{\mathbb{E} \left[(X - \mu)^n \right]}{\sigma^n} 
\end{equation}
To demonstrate the usefulness of higher moments aggregation and further motivate the need for multiple aggregation functions, we ran an ablation study showing how different moments affect the performance of the model. We conduct this by testing five different models, each taking a different number of moments, on our multi-task benchmark.

\begin{figure}[h]
\centering
\includegraphics[width=11cm]{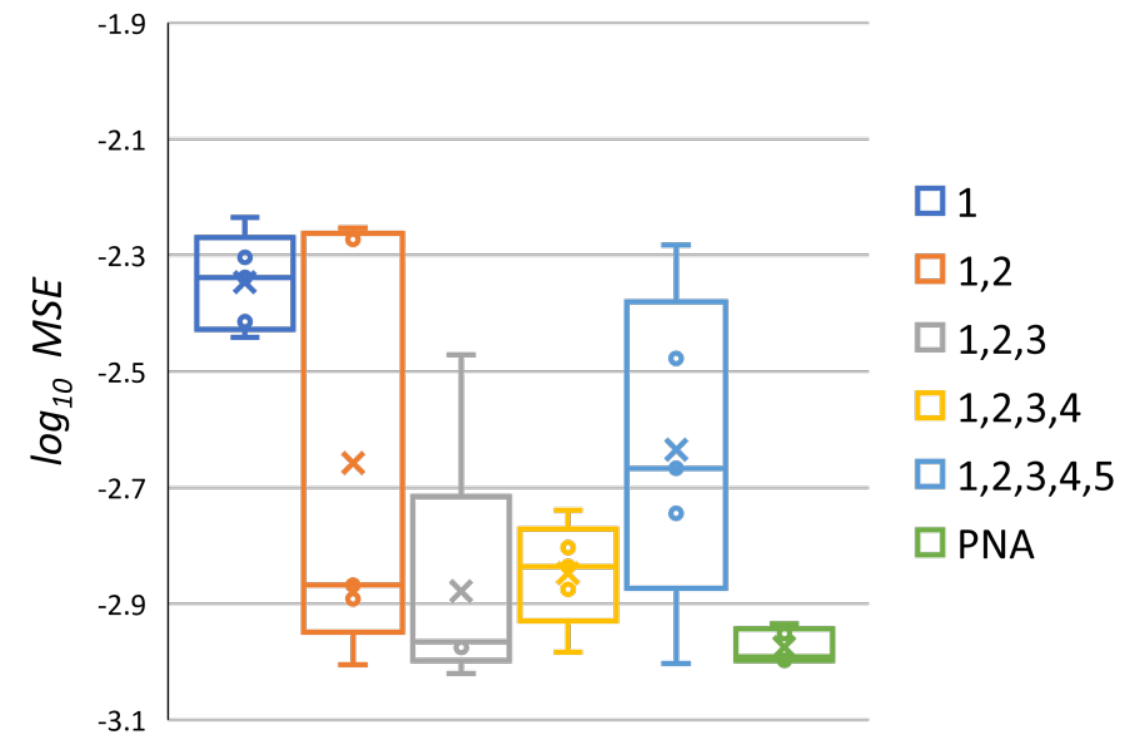}
\vspace{15pt}
\caption{Multi-task $\log_{10}$ MSE on different versions of the PNA model with increasing number of moments aggregators (specified in the legend), using \textit{mean} as first moment. All the models use the identity, amplification and attenuation scalers. The model on the right is the complete PNA as described before (\textit{mean}, \textit{max}, \textit{min} and \textit{std} aggregators).}
\label{fig:moments_plot}
\end{figure} 

The results in Figure \ref{fig:moments_plot} demonstrate that with the increase of the number of aggregators the models reach a higher expressive power, but at a certain point (dependent on the graphs and tasks, in this case around 3) the increase in expressiveness given by higher moments reduces the performance since the model becomes harder to optimize and prone to overfitting. We expect that higher moments will be more beneficial on graphs with a higher average degree since they will better characterize the neighbourhood distributions. 

Finally, we note how the addition of the \textit{max} and \textit{min} aggregators in the PNA (rightmost column) gives a better and more consistent performance in these tasks than higher moments. We believe this is task-dependent, and, for algorithmic tasks, discrete aggregators can be valuable. As a side note, we point out how the \textit{max} and \textit{min} aggregators of positive values can be considered as the n\textsuperscript{th}-root of the n\textsuperscript{th} (non-centralized) moment as n tends to, respectively, $+\infty$ and $-\infty$.

\pagebreak

\section{Alternative aggregators}
\label{app:aggregators}

Besides those described above, we have experimented with additional aggregators. We detail some examples below. Domain-specific metrics can also be an effective choice.

\paragraph{Softmax and softmin aggregations}
As an alternative to \textit{max} and \textit{min}, \textit{softmax} and \textit{softmin} are differentiable and can be weighted in the case of edge features or attention networks. They also allow an asymmetric message passing in the direction of the strongest signal. Equation \ref{eq:softmax} presents their direct neighbour formulations, where $X^l$ are the nodes features at layer $l$ with respect to node $i$ and $N(i)$ is the neighbourhood of node $i$:

\begin{equation}
    \label{eq:softmax}
    \text{softmax}_i(X^l) = \sum_{j \in N(i)} 
    \frac{X^l_j \: \exp(X^l_j)}{\sum_{k \in N(i)} \exp(X^l_k)}
    \quad,\quad
    \text{softmin}_i(X^l) = - \text{softmax}_i(-X^l)
\end{equation}

\section{Alternative graph convolutions}
\label{app:convolutions}

In this section, we present the details of the four graph convolutional layers from existing models that we used to compare the performance of the PNA in the multi-task benchmark.

\paragraph{Graph Convolutional Networks (GCN)} \cite{kipf2016gcn} use a normalized mean aggregator followed by a linear transformation and an activation function. We define it in Equation \ref{eq:gcn}, where $\tilde{A} = A + I_N$ is the adjacency matrix with self-connections, $W$ is a trainable weight matrix and $b$ a learnable bias.
\begin{equation}
    \label{eq:gcn}
    X^{(t+1)} = \sigma \left( \tilde{D}^{-\frac{1}{2}} \tilde{A} \tilde{D}^{-\frac{1}{2}} X^{(t)} W + b \right)
\end{equation}

\paragraph{Graph Attention Networks (GAT)} \cite{velikovic2017gat} perform a linear transformation of the input features followed by an aggregation of the neighbourhood as a weighted sum of the transformed features, where the weights are set by an attention mechanism $a$. We define it in Equation \ref{eq:gat}, where $W$ is a trainable projection matrix. As in the original paper, we employ the use of multi-head attention.
\begin{equation}
    \label{eq:gat}
    X^{(t+1)}_i = \sigma \left( \sum_{(j,i) \in E} a \left( X^{(t)}_i, X^{(t)}_j \right) W X^{(t)}_j \right)
\end{equation}

\paragraph{Graph Isomorphism Networks (GIN)} \cite{xu2018gin} perform a sum aggregation over the neighbourhood, followed by an update function $U$ consisting of a multi-layer perceptron. We define it in Equation \ref{eq:gin}, where $\epsilon$ is a learnable parameter. As in the original paper, we use a 2-layer MLP for $U$.
\begin{equation}
    \label{eq:gin}
    X^{(t+1)}_i = U \Bigg( \Big( 1 + \epsilon \Big) X^{(t)}_i + \sum_{j \in N(i)} X^{(t)}_j \Bigg)
\end{equation}

\paragraph{Message Passing Neural Networks (MPNN)} \cite{gilmer2017mpnn}  perform a transformation before and after an arbitrary aggregator. We define it in Equation \ref{eq:mpnn}, where $M$ and $U$ are neural networks and $\bigoplus$ is a single aggregator. In particular, we test models with \textit{sum} and \textit{max} aggregators, as they are the most used in literature. As with PNA layers, we found that linear transformations are sufficient for $M$ and $U$ and, as in the original paper \cite{gilmer2017mpnn}, we employ multiple towers.
\begin{equation}
    \label{eq:mpnn}
    X_i^{(t+1)} = 
    U \Bigg( X_i^{(t)}, 
    \underset{(j,i) \in E}{\bigoplus} 
    M \Big( X_i^{(t)}, X_j^{(t)} \Big) \Bigg)
\end{equation}

\section{Random graph generation}
\label{app:graph_generation}

In this section, we present the details of the random generation of the graphs we used in the multi-task benchmark. Following previous work \cite{velickovic2019neural,you2019positionaware}, we opted for undirected unweighted graphs from a wide variety of types (we provide, in parentheses, the approximate proportion of such graphs in the benchmark). Letting $N$ be the total number of nodes per graph:

\begin{itemize}
    \item \textbf{Erd\H{o}s-R\'{e}nyi} \cite{erdos1960} (20\%): with probability of presence for each edge equal to $p$, where $p$ is independently generated for each graph from $\mathcal{U}[0,1]$
    
    \item \textbf{Barab\'{a}si-Albert} \cite{albert2002} (20\%): the number of edges for a new node is $k$, which is taken randomly from $\{1, 2, ..., N-1 \}$ for each graph
    
    \item \textbf{Grid} (5\%): $m \times k$ 2d grid graph with $N = mk$ and $m$ and $k$ as close as possible 
    
    \item \textbf{Caveman} \cite{watts1999caveman} (5\%): with $m$ cliques of size $k$, with $m$ and $k$ as close as possible 
    
    \item \textbf{Tree} (15\%): generated with a power-law degree distribution with exponent 3 
    
    \item \textbf{Ladder graphs} (5\%)
    
    \item \textbf{Line graphs} (5\%)
    
    \item \textbf{Star graphs} (5\%)
    
    \item \textbf{Caterpillar graphs} (10\%): with a backbone of size $b$ (drawn from $\mathcal{U}[1,N)\,$), and $N-b$ pendent vertices uniformly connected to the backbone 
    
    \item \textbf{Lobster graphs} (10\%): with a backbone of size $b$ (drawn from $\mathcal{U}[1,N)\,$), $p$ (drawn from $\mathcal{U}[1,N-b \, ]\,$) pendent vertices uniformly connected to the backbone, and additional $N-b-p$ pendent vertices uniformly connected to the previous pendent vertices.
\end{itemize}

Additional randomness was introduced to the generated graphs by randomly toggling arcs, without strongly impacting the average degree and main structure. If $e$ is the number of edges and $m$ the number of 'missing edges' ($2e + 2m = N(N-1)$), the probabilities of toggling an existing and missing edge, respectively $P_e$ and $P_m$, are:
\begin{equation}
P_{e} =
\begin{cases}
    \begin{array}{ll}
      0.1 & e \leq m\\
      0.1 \; \frac{m}{e} & e > m
    \end{array}
\end{cases}
\quad P_{m} =
\begin{cases}
    \begin{array}{ll}
      0.1 \; \frac{e}{m} & e \leq m\\
      0.1 & e > m
    \end{array}
\end{cases}
\end{equation}
After performing the random toggling, we discarded graphs containing singleton nodes, as they are in no way affected by the choice of aggregation.

\section{Graph type experiments} \label{app:graph_type}

In order to better interpret the improvements in performance that the PNA brings, we tested the models against the various types of graphs in the multi-task benchmark. In particular, in these experiments, we trained the models on the whole dataset with the proportions described above and then tested them against datasets composed by just one category of graphs. 

\begin{figure}[h]
\centering
\includegraphics[width=0.95\textwidth]{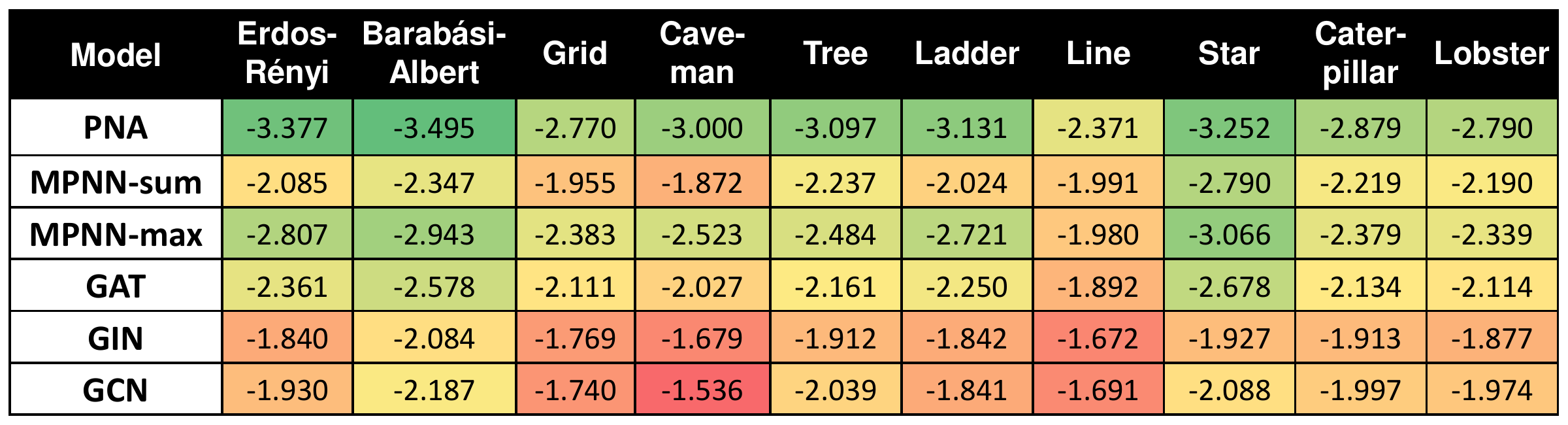}
\caption{Average $\log_{10}$MSE error across the various tasks of a particular model against a particular type of graphs.}
\label{fig:graph_type}
\end{figure}

The results, presented in Figure \ref{fig:graph_type}, show that the PNA improves across all types. However, it performs the worst on the graphs with a higher diameter (especially graphs close to lines), suggesting that the number of layers is not enough to reach the complete graph. Therefore, the main limitation to the PNA performance seems to be the message passing framework; this could motivate future research to try to improve the framework itself.

\section{Standard architecture} \label{app:std_architecture}

In this section we will provide more intuition on the motivation behind our choice of architecture, presented in Section \ref{sec:architecture}, which we will refer to as \textit{recurrent},\footnote{Note that this was only used in the synthetic benchmarks, while in the real-world benchmarks, we kept the same architecture from Dwivedi \textit{et al.}} and present the results on a more \textit{standard} architecture.

The main motivations behind the choice of the architecture were: (1) provide a fairer comparison between the models (2) showcase a parameter-efficient \textit{recurrent} architecture with a prior \footnote{This prior corresponds to the knowledge that these tasks can be solved by the convergence of an aggregation function in the message passing context, potentially with an additional readout/function.} that works very well with the tasks at hand. In particular:
\begin{enumerate}
    \item The GRU helps to avoid over-smoothing, and the models that do not have a skip connection across the aggregation (GAT, GIN and GCN) are those benefiting the most from it; therefore, to still provide a fair comparison in the results below, we added  skip connections from every convolutional layer to the readout, in all the models. 
    \item The S2S (as opposed to a mean readout used below) helps the most architectures without scalers as it can provide an alternative counting mechanism. 
    \item The repeated convolutions are a parameter-saving prior which works well in these tasks but does not change the rank between the various models.
\end{enumerate}

\begin{figure}[h]
\centering
\includegraphics[width=0.8\textwidth]{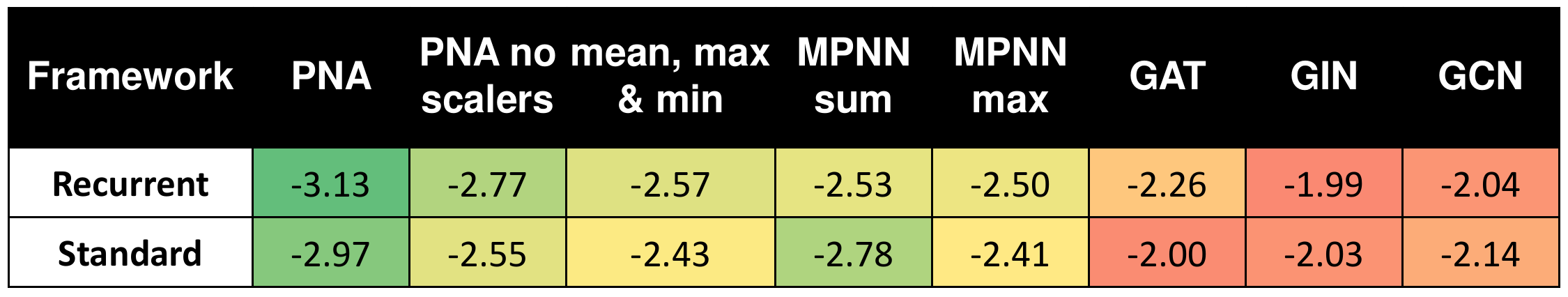}
\caption{Average $\log_{10}$MSE error across the various tasks of a particular model when inserted in the \textit{recurrent}  or the \textit{standard} model. The \textit{mean, max \& min} model represents a baseline MPNN which employs \textit{mean}, \textit{max} and \textit{min} aggregators and no scaler.}
\label{fig:standard_architecture}
\end{figure}

For completeness, we present in Figure \ref{fig:standard_architecture} the comparison of the average results of the \textit{recurrent} architecture and \textit{standard} one which uses no GRU but skip connections, mean readout rather than S2S and a fixed number of convolutions (8).

\section{Single task experiments} \label{app:single_task}

\begin{figure}[h]
\centering
\includegraphics[width=0.7\textwidth]{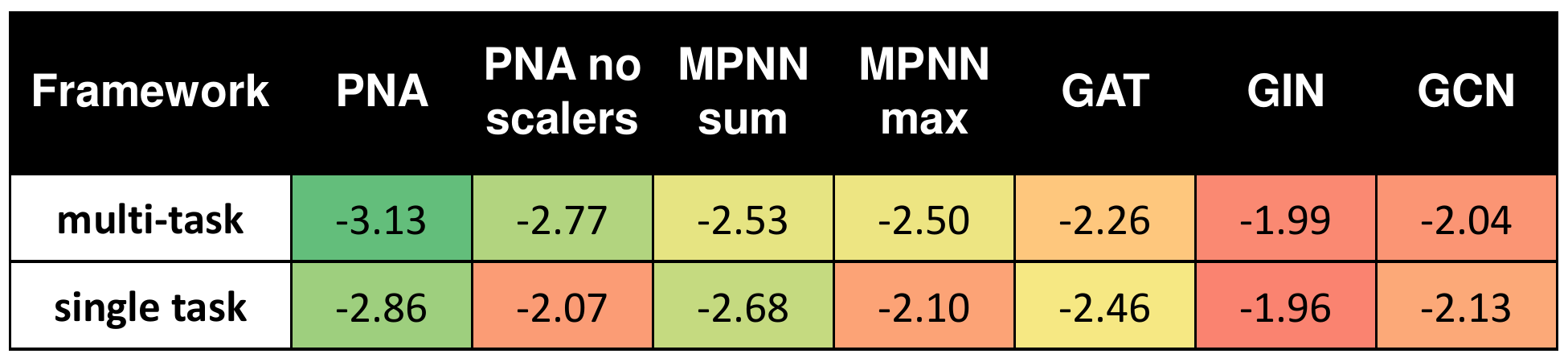}
\caption{Average $\log_{10}$MSE error across the various tasks of a particular model either trained concurrently on all the tasks (\textit{multi-task}) or trained separately on the individual tasks (\textit{single task}). With the exception of the output layer, the two settings use the same architecture.}
\label{fig:single_task}
\end{figure}

Apart from a good method to evaluate the performance on a variety of different problems, the multi-task approach offers a regularization opportunity that some models capture more than others. In particular, we found that models without scalers (or \textit{sum} aggregator) are those benefiting the most from the approach; we hypothesise that the reason for this lies in some supervision that specific tasks give to recognise the size of a model neighbourhood. Moreover, more complex models are more prone to overfitting when trained on a single task. Figure \ref{fig:single_task} shows the average performance on the individual tasks of the various models.

\section{Parameters comparison}
\label{app:parameters}

\begin{figure}[h]
\centering
\includegraphics[width=0.5\textwidth]{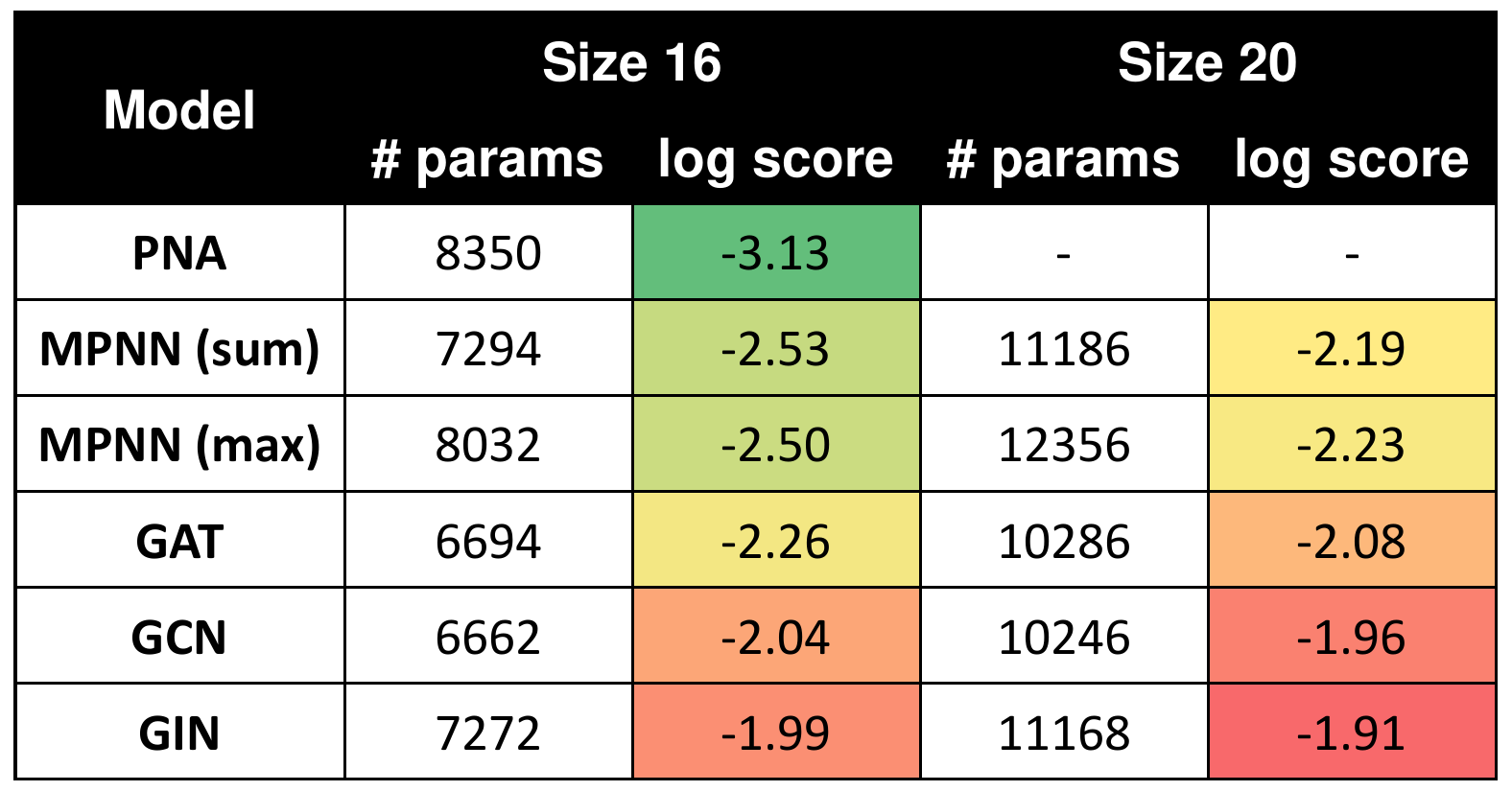}
\caption{Average score of different models using feature sizes of 16 and 20, compared to the PNA with 16 on the multi-task benchmark.  "\# params" is the total number of parameters in each architecture. }
\label{fig:number_params}
\end{figure}

Figure \ref{fig:number_params} shows the results of testing all the other models on the multi-task benchmark with increased latent size.

We observe that, even with fewer parameters, PNA performs consistently better and an increased number of parameters does not boost the performance of the other models. This suggests that the multiple aggregators in the PNA produce a qualitative improvement to the capacity of the model.

\end{document}